\documentclass{article}

\usepackage{arxiv}
\usepackage[utf8]{inputenc} 
\usepackage[T1]{fontenc}    
\usepackage{hyperref}       
\usepackage{url}            
\usepackage{booktabs}       
\usepackage{amsfonts}       
\usepackage{lipsum}
\usepackage{graphicx}
\usepackage{xcolor}
\usepackage{float}

\title{Waste detection in Pomerania: non-profit project for detecting waste in environment}

\author{
  Sylwia Majchrowska \\
  Wroclaw University of Science and Technology\\
  Wrocław, Poland \\
  \texttt{sylwia.majchrowska@pwr.edu.pl} \\
   \And
 Agnieszka Mikołajczyk \\
  Gdańsk University of Technology\\
  Gdańsk, Poland \\
  \texttt{agnieszka.mikolajczyk@pg.edu.pl} \\
  \And
 Maria Ferlin \\
  Gdańsk University of Technology\\
  Gdańsk, Poland \\
  \texttt{maria.ferlin@pg.edu.pl} \\
  \And
 Zuzanna Klawikowska \\
  Gdańsk University of Technology\\
  Gdańsk, Poland \\
  \texttt{zuwikowska@gmail.com} \\
  \And
 Marta A. Plantykow \\
  Intel Corporation \\
  Gdańsk, Poland \\
  \texttt{m.plantykow@gmail.com} \\
  \And
 Arkadiusz Kwasigroch \\
  Gdańsk University of Technology\\
  Gdańsk, Poland \\
  \texttt{arkadiusz.kwasigroch@pg.edu.pl} \\
  \And
 Karol Majek \\
  Cufix\\
  Warsaw, Poland \\
  \texttt{karolmajek@cufix.pl}
}

\begin{document}
\maketitle

\begin{abstract}
Waste pollution is one of the most significant environmental issues in the modern world. The importance of recycling is well known, either for economic or ecological reasons, and the industry demands high efficiency. Our team conducted comprehensive research on Artificial Intelligence usage in waste detection and classification to fight the world's waste pollution problem. As a result an open-source framework that enables the detection and classification of litter was developed. The final pipeline consists of two neural networks: one that detects litter and a second responsible for litter classification. Waste is classified into seven categories: \textit{bio}, \textit{glass}, \textit{metal and plastic}, \textit{non-recyclable}, \textit{other}, \textit{paper} and \textit{unknown}. Our approach achieves up to 70\% of average precision in waste detection and around 75\% of classification accuracy on the test dataset. The code used in the studies is publicly available online\footnote{\url{https://github.com/wimlds-trojmiasto/detect-waste}}.
\end{abstract}


\section{Introduction}
\label{sec:intro}
Litter is scattered in a wide variety of environments. The massive production of disposable goods in the last years resulted in an exponential increase in produced garbage, reaching about 2 billion tons per year~\cite{WB16}. It was estimated that humans eat up to 250 grams of micro plastics each year~\cite{NC19}. Almost 90 percent of this plastic comes from bottled and tap water. Today, more than 300 million tons of plastic is produced annually and only 30 years are left for the amount of garbage in the ocean to exceed the number of sea creatures~\cite{WE16}.

The most difficult challenge in the waste management is complex, and not unified guidelines regarding segregation rules. Due to the large energy requirements and related costs plastic waste separation on many litter sorting lines is done manually. However, new possibilities for the automatic selection of these materials and their reuse are constantly being tested~\cite{plastic_sorting}.

In recent years machine learning (ML) based systems that can support or fully cover sorting processes were implemented, accelerating this procedure as a result. The most commonly used solutions include self-sorting smart bins, which are capable of classifying one object located on a clear background at a time~\cite{white2020wastenet,Sheng2020}. When a single compartment is used, the camera is usually located at the top of the upper container.The deep learning (DL) model assigns proper class based on the photo and the garbage is moved to the appropriate bottom container~\cite{white2020wastenet}. Another way is to a mount camera or a sensor above few separate bins, and direct the consumer handling waste to the correct one~\cite{IoWT2018}. In this approach, the rubbish must be well exposed to the imaging device. Here it comes to the fact that the problem of identifying garbage in the wild environment is ambiguous. An image could present different types of trash, which additionally can be deformed, and/or recorded in a uncontrolled various natural scenarios. On the other hand, as the same object may be a junk or not depending on the context, it must be defined at what point localized object should be treated as a litter.

This paper addresses some of the identified limitations. At first, to provide more efficient recycling, we mixed publicity available datasets of waste observed in different environments, and proposed seven well-defined categories for sorting litter: \textit{bio, glass, metal and plastic, non-recyclable, paper, other} and \textit{unknown}. Their selection was inspired by rules in Gdansk (Poland). Secondly, we implemented a two-stage DL-based framework for waste detection that consisted of two separate neural networks: detector and classifier. The proposed framework is freely available and can be used for different purposes, such as studying the common types of waste observed in nature. To our knowledge, we presented the first experiments that allow for such universal litter detection and classification. Additionally we provided the first comprehensive review of existing waste datasets.

An overview of the actual work for deep learning classification and detection, as well as existing public datasets of waste are described in Section~\ref{sec:related_works}. Section~\ref{sec:our_approach} illustrates our framework to detect and classify waste from images, and statistics of examined data. More specifically, we described used data in details, provide the training details of chosen neural networks, and reported the obtained results. Finally, conclusions are drawn and future work is outlined in Section~\ref{sec:conclusions}.

\section{Related works}
\label{sec:related_works}

ML and DL techniques empower many aspects of modern society, like recommendation systems, text-to-speech devices, or even objects identification in images~\cite{lecun2015deep}. Also the waste management problem has attracted a lot of interest~\cite{white2020wastenet,Sheng2020,Glouche2013ASW}, where the main goal is to create an ML-based image recognition system to sort litter. The majority of the proposed approaches are based on the deep learning algorithms utilized in the computer vision field. This section describes chosen techniques used in classification and object detection challenges and delivered a comprehensive review of existing public waste datasets.

\subsection{Classification}
\label{sec:classification}
Convolutional neural networks (CNNs) have had a massive impact on large-scale image classification tasks and made it possible to achieve significantly higher accuracy than solutions based on classical image processing. Multiple convolutional layers stacked together can automatically learn representations from the data reducing the need for manual feature extraction engineering~\cite{lecun2015deep}. The design of neural network structure impacts the performance, latency, and computational requirements. Thus efforts have shifted from feature extraction design to architecture design. Architectural enhancements of deep neural networks improve the performance of classification, segmentation and detection tasks.

A breakthrough in CNN performance came with AlexNet \cite{NIPS2012_4824}, a neural network that won the ImageNet Large Scale Visual Recognition 2012 challenge. Since then, the quality of image recognition structures has advanced rapidly. AlexNet consists of a stack of five convolutional layers of different kernel sizes. The structure incorporates non-saturating ReLU activation functions, overlapping pooling, and normalization layers. Dropout layers were used to reduce overfitting. %

The authors of \cite{DBLP:journals/corr/SimonyanZ14a} proposed a family of architectures called VGG - very deep networks consisting of 16 to 19 layers. The authors observed that stacking of several 3x3 layers could imitate the kernels of bigger sizes achieving greater effective receptive fields while reducing the number of parameters. This strategy provides an additional benefit of a more discriminative decision function caused by more nonlinear layers in the structure.

Further increase in the number of layers leads to deterioration of training and decrease in neural network performance. To alleviate the problem ResNet family \cite{resnet50} was proposed. The architecture is based on residual connections, that introduce no extra parameters or computational effort. Instead, this connectivity pattern facilitates gradient flow, enabling effective training of networks consisting of as many as 200 layers.

Inception \cite{43022} network stacks modules called Inception instead of single layers. Inception blocks incorporate multi-scale convolutional transformations utilizing a split-transform-merge strategy. Moreover, it allows for increasing the number of units at each stage without an uncontrolled blow-up in computational complexity. The Inception models have evolved by utilizing batch normalization and factorization of convolutional layers \cite{44903}, simplification, and the use of asymmetric filters \cite{10.5555/3298023.3298188}. The proposed Inception-Resnet structure \cite{10.5555/3298023.3298188} combines both the power of inception modules and residual connections.

The improvement of connectivity pattern in the DenseNet structure \cite{8099726} facilitated the training and accuracy. For each layer, the feature maps of all preceding layers are used as inputs, and its feature maps are used as inputs into all subsequent layers. DenseNets have several significant advantages: they minimize the vanishing gradient problem, improve feature propagation, encourage feature reuse, and reduce the number of parameters.

ResNeXt \cite{8100117} structure similarly to Inception utilizes a split-transform-merge strategy. However, unlike all Inception or Inception-ResNet modules, the same topology among the multiple paths is shared and paths are aggregated by summation. The authors introduced a conception of cardinality - the number of paths in one module. It was observed that an increase in cardinality is more effective than going deeper or wider when increasing the capacity.

EfficientNet \cite{efficientnet2019} architecture consists of modules constructed by the neural architecture search process that optimizes both for accuracy and FLOPS. The mobile-size baseline model called EfficientNet-B0 was built stacking those modules. Scaling strategies were used to produce more complex and accurate models EfficientNet-B1-7. Scaling of a convolutional neural network most often is performed in one of the following dimensions width, depth, or resolution. Whereas, EfficientNets scale three dimensions jointly, significantly improving efficiency and accuracy

EfficientNetv2 \cite{tan2021efficientnetv2} improves EfficientNet by providing faster training and better parameter efficiency by eliminating EfficientNet bottlenecks. Neural architecture search was used to optimize accuracy, training speed, and parameter size. Unlike, standard EfficientNet, EfficientNetv2 uses non-uniform scaling of depth, resolution, and width. Moreover, to limit computational cost an increase in resolution was limited.

Inspired by the successes in Natural Language Processing, recent computer vision architecture advances rely on the Transformer module. The authors of \cite{dosovitskiy2020image} trained Visual Transformer (Vit) --  a pure transformer architecture applied directly to sequences of image patches that perform very well on image classification tasks.
The authors stated that training transformers for vision tasks requires large amounts of training samples and extensive computing resources.
This particular trait was addressed in Data-efficient image Transformers
(DeiT) \cite{touvron2020training} in which a new model distillation procedure is shown.
In \cite{wang2021pyramid} authors present Pyramid Vision Transformer (PVT) which is designed for dense prediction tasks such as object detection.
PVT introduces to transformers the pyramid structure known from CNNs.
Therefore it is more suitable for object detection tasks than Vit, which was designed for image classification.
The application of transformers in vision is further investigated in \cite{wu2021cvt}.
Authors introduce convolutions to ViT architecture to introduce shift, scale,
and distortion invariance.
Proposed Convolutional vision Transformer (CvT) architecture accomplishes this goal by two main modifications: introduction of a hierarchy of Transformers containing a new convolutional token embedding, and a convolutional Transformer
block leveraging a convolutional projection.
The application of transformers in vision tasks is not yet a well-investigated topic, and we observe many teams that are currently working on improving it.

\subsection{Object Detection}
\label{sec:detection}
Object detection is a well-studied task in computer vision \cite{zou2019object}.
It is defined as a localization of Axis-Aligned Bounding Box (AABB) and classification - assignment of a single or multi-label.
In many previous works object detection was approached using two types of techniques namely one-stage and two-stage detection.
One-stage architectures provide both locations and classes for each object in a single step, while two-stage detectors find class-agnostic object proposals first, classifies them into the class-specific detections in the second stage.

\textbf{Two-stage detectors} were the first object detection methods.
They used the sliding window approach in the image pyramid to generate object proposals in multiple scales.
Then in a second stage a classifier such as a cascade classifier \cite{viola2001rapid} was used.
Such a system achieved 15 Hz for the face detection system which is a single class object detection problem.
A significant improvement was presented in \cite{dalal2005histograms} where authors introduced a novel descriptor namely Histograms of Oriented Gradient (HOG) and showed a method of human detection using Support Vector Machine (SVM) to classify objects.
The authors used a detection window to scan across the image at all positions and scales.
Classification is invoked multiple times in similar regions and using multiple image pyramid levels.

Using convolutional neural networks for object detection was a challenging task since CNNs' inability to localize features.
In most approaches, this problem was solved by using the \textit{recognition using regions} paradigm \cite{gu2009recognition}.
Identified regions can yield richer information than single pixels or pixels with a fixed local neighbourhood.
Regions are used for object detection successfully in the Selective Search algorithm \cite{uijlings2013selective}.
Authors generate initial regions which are grouped by similarity and merged in the iterative process.
Object hypotheses from regions were used to extract features using key point feature descriptors and classify those proposals using SVM.
Selective Search was also used in R-CNN \cite{girshick2014rich} to extract region proposals, on which CNN features where computed to classify regions using per class SVMs finally.
Fast R-CNN \cite{girshick2015fast} solved R-CNN's main problems: multiple training sessions required since SVMs are involved, inefficient processing of regions through the CNN.
Fast R-CNN computes convolutional features for the whole image in the first step to reduce computations for overlapping regions.
The other improvement is integrating classification  into the network architecture, which allows allows training of the entire
network in a single multi-task training session.
When Faster R-CNN \cite{ren2015faster} was introduced State-of-the-art object detection networks used region proposal algorithms to generate object location hypotheses.
In this architecture a Region Proposal Network (RPN) concept was  used instead of the Selective Search algorithm.
RPN is integrated into Fast R-CNN architecture and uses shared weights.
Many researchers used Faster R-CNN while replacing its backbone (feature extraction CNN) with newer architectures in the following years. Faster R-CNN is like its' predecessors, a two-stage object detection method.

\textbf{Single-stage} detection was popularized in Deep Learning mainly by two detector architectures: Single Shot MultiBox Detector (SSD) \cite{liu2016ssd} and You Only Look Once (YOLO v1, 9000, v3, v4, scaled V4: \cite{redmon2016you,redmon2017yolo9000,redmon2018yolov3,bochkovskiy2020yolov4,wang2020scaled}).
Those networks perform both, object detection and object classification in a single step. Therefore the detection time can be reduced.
Such a result is obtained by generating a constant number of predictions per class per image.
For SSD it was 8732 detections for each class for a 300x300px image.
Each detection is characterized by a category score for each category and 4 offsets for a fixed set of default bounding boxes.
Default boxes have different aspect ratios at each location in several feature maps with different scales.
In this type of methods, many false detections need to be removed considering objectness or classification score and overlaps between detections.
YOLOv1 used a single scale image for prediction with 98 detections per class and did not use default boxes
YOLOv2 (YOLO9000) predicts nine bounding boxes at each cell from a 13x13 grid, resulting in in 1521 boxes.
YOLOv3 introduces a multi-label approach to classification since, in datasets such as Open Images \cite{kuznetsova2020open}, class labels do not have a guaranteed hierarchy level.
It uses image 3 image pyramid levels for detection.
YOLOv4 introduces a new backbone network while using the same detector as YOLOv3.
This paper studies data augmentation, post-processing methods.
A month after YOLOv4 \cite{bochkovskiy2020yolov4} was proposed, a new approach to object detection tasks was introduced in \cite{carion2020end}.
Detection Transformer (DETR) introduced Transformers known from Natural Language Processing tasks \cite{vaswani2017attention} into object detection while preserving image processing with CNNs.
DETR predicts a sparse, fixed number of objects which are in training matched with ground truth labels using bipartite matching.
DETR's main limitations were slow convergence and limited feature spatial resolution.
It required 10 times more epochs than other approaches to converge to a similar error level.
Limited spatial resolution lowered the Average Precision metric for small objects' sizes.
The following year, the main issues were mitigated in Deformable DETR \cite{zhu2020deformable} thanks to the introduced deformable multi-head attention module.

EfficientDet, a single-stage object detector, was introduced in \cite{efficientdet2020} a month after \cite{carion2020end}.
The main goal of the authors was to improve the efficiency of object detection models.
The goal was separated into two challenges, namely efficient multi-scale feature fusion and model scaling.
They introduced a weighted bi-directional feature pyramid network (BiFPN) to address the first challenge.
BiFPN adds a bottom-up pathway to fuse multi-scale features similarly to Neural Architecture Search Feature Pyramid Network (NAS-FPN) \cite{ghiasi2019fpn}, which used NAS to find the best top-down, bottom-up pathways for feature fusion.
The second challenge was addressed using a compound scaling method for object detectors inspired by \cite{efficientnet2019}.
It jointly scales up the resolution/depth/width for all backbone, feature network, and box/class prediction network.
EfficientDet is proposed in 8 variants D0-D7 with backbone networks EfficientNet B0-B7.
The approach shown in \cite{efficientdet2020} reduced the number of parameters and latency in the object detection network and simultaneously increased the Average Precision metrics.

\textbf{Instance Segmentation} is another task for automated image processing.
The goal of this task is to provide a segmentation mask for each object instance.
Approaches introduced in the \textit{pre-Deep Learning Era} relied on methods such as edges or superpixels.
This changed in Deep Mask \cite{DBLP:journals/corr/PinheiroCD15} which was one of the first deep neural networks used for the instance segmentation task.
The proposed network predicts object proposals for the whole image.
Those proposals are class-agnostic segmentation masks, and for each of them, an object likelihood score is computed.
Mask RCNN \cite{he2017mask} is an extension of Faster R-CNN, and it ads the instance mask branch in parallel to the classification and bounding box regression branch.
In Mask RCNN recognition precedes segmentation, which is faster and more accurate according to authors.

\subsection{Data collections}
\label{sec:litter_datasets}
In recent years multiple attempts to detect, classify and segment waste using deep learning have been made. Litter classification of common waste categories based on images has been attempted using few pre-trained convolutional neural networks -- AlexNet, MobileNet, InceptionResNetV2, DenseNet, Xception~\cite{TrashNet2016, Bircanoglu18, Chu2018} -- achieving average accuracy in ranges 22\%~\cite{TrashNet2016} and 98.2\%~\cite{Chu2018} for pictures of waste on a plain background. Experiments have also been conducted on the detection of litter on the streets and homes, as well as the segmentation of different types of waste using: Faster R-CNN, Mask R-CNN, SSD, and different types of YOLO (even Tiny-YOLO)~\cite{Awe2017trashnet, Liu2018, ICRA2019, hong2020trashcan, proencca2020taco, YOLO-TrashNet2020, MJU-Waste2020, UAVVaste2021}. Calculated mAP value for detection task varies between different datasets and architectures from 15.9\% for TACO~\cite{proencca2020taco} with Mask-RCNN, up to 81\% for Trash-ICRA19~\cite{ICRA2019} with Faster R-CNN.

\begin{figure}[!hbt]
\centering
  \includegraphics[width=0.7\textwidth]{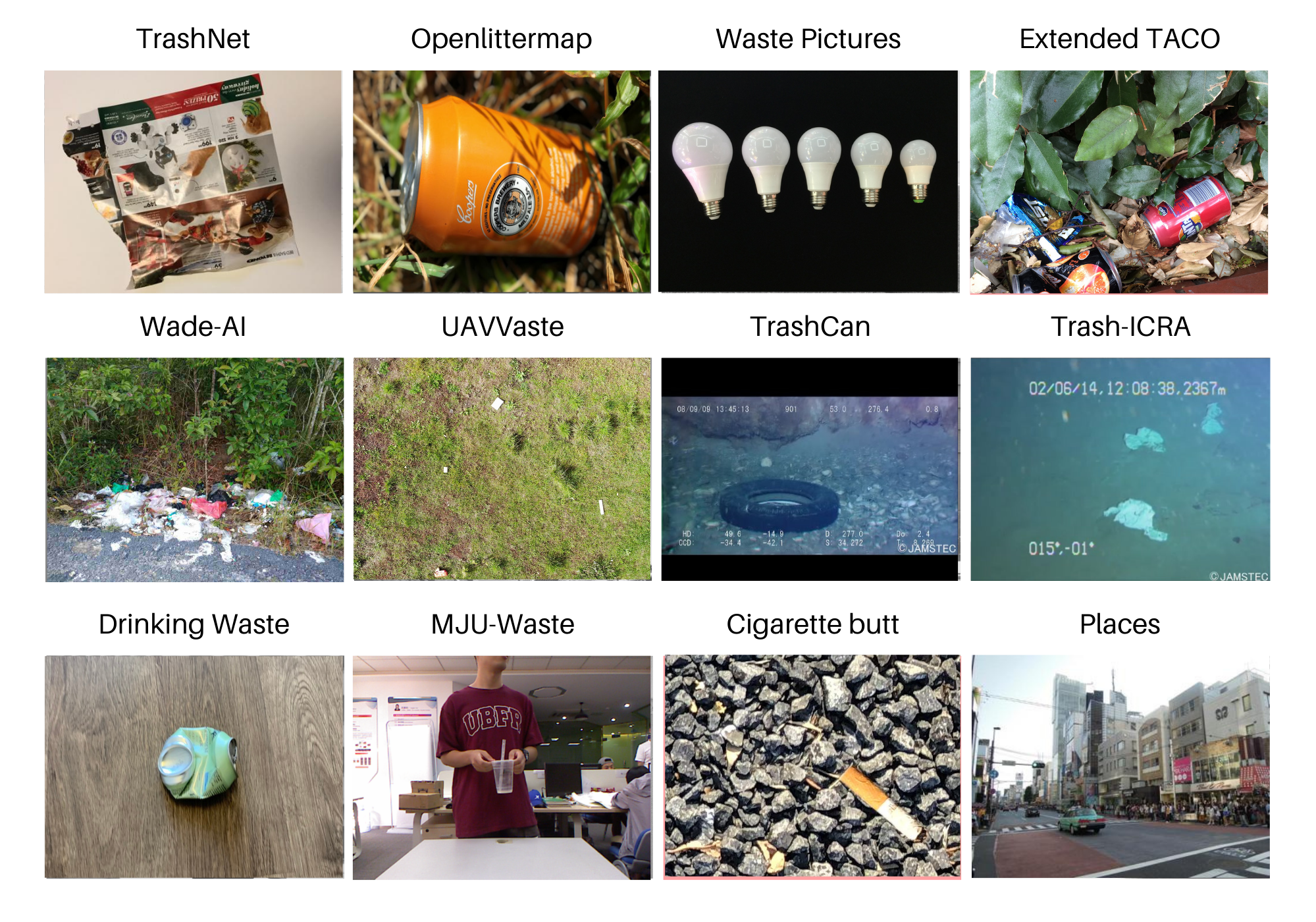}%
  \caption{Representative pictures from each dataset. \label{fig.dataset_representants}}
\end{figure}

Nonetheless, very few attempts at the detection and classification into well-recognized recyclable classes have been made. As deep learning requires diverse data to fit the model's parameters optimally, a comprehensive review of existing litter datasets in terms of litter detection is provided. Over ten different data collections and their crucial statistics (e.g. number of images, annotation type) are presented in Table~\ref{tab:dataset_stats} (see Fig~\ref{fig.dataset_representants} to visualize their representatives).

\textbf{TrashNet.} The TrashNet dataset~\cite{TrashNet2016} contains over 2100 labeled images. Each of them belongs to one of the six classes: glass, paper, cardboard, plastic, metal, and trash. The pictures were taken by mobile phone camera using sunlight and/or room lighting. Photographed objects were placed on a white background or fulfill the whole view (cardboard). All images have size a of 512x384px.

\textbf{Open Litter Map.} Open Litter Map~\cite{openlittermap2018} is a free, open, and crowd-sourced dataset with over 100k images taken by phone cameras. All images are provided with information such as type of presented litter, coordinates, timestamp or phone model. Images come from all over the world, taken by different people. Therefore, they differ significantly from one another.

\textbf{Waste Pictures.} Waste Pictures~\cite{waste_pictures2019} contains almost 24000 waste images scraped from Google search, divided into 34 classes.  The type of images is very diverse, including even x-rays and drawings of garbage. Sizes also differ significantly. However, most of the photos are below the size of 2000x2000px. Due to the origin of images, they should be manually reviewed for use in a classification task.

\textbf{Extended TACO.} Trash Annotations in Context (TACO) ~\cite{proencca2020taco} is a crowd-sourced dataset of waste in the wild with high-resolution mobile phone images. The TACO dataset contains 1500 annotated images with almost 5000 objects. All trash have been assigned to one of 60 classes that belong to 28 super (top) categories, including the category \textit{Unlabeled litter} for hard to recognize or heavily obscured objects. The annotations are provided in the well-known COCO format~\cite{bib:COCO2014} on the instance segmentation level with an extra background description - Trash, Vegetation, Sand, Water, Indoor, Pavement. Additionally, TACO offers around three thousand of unannotated images, which we have taken advantage of: we provided annotations on the detection level\footnote{Only the primary part (1.5k images) of the dataset is annotated with instance masks, we provided bounding box annotation for the rest.} for over 3000 images achieving over 14~000 instances in total. A great advantage is that TACO is characterized by various litter types and high diversity of the backgrounds, from tropical beaches to London streets. However, due to the crowd-sourcing nature of the dataset, labels may contain some user-induced errors and bias, i.e., not all objects in TACO can be categorized strictly as litter as their category is often based on context.

\textbf{Wade-AI.} The Wade-AI dataset~\cite{wade-ai2016} contains images of waste in the wild environment, provided by Google Street View. It consists of nearly 1400 images with 2200 manually labeled instance masks annotations in COCO format with only one class, called rubbish. The environment and size of the images vary due to the source of the images. Most images are less than 1000x1000.

\textbf{UAVVaste.} Another  publicly available dataset, which also provides instance segmentation masks in the COCO format, is UAVVaste~\cite{UAVVaste2021} dataset. It contains 772 hand-labeled aerial images of waste with over 3700 objects of one class - "rubbish". Data was collected in the urban and natural environments e.g. streets, parks and lawns using Unmanned Aerial Vehicles (UAV). The annotated litter is usually relatively small (median of object shape is 76x68px, while median of image shape is 3840x2160px).

\textbf{TrashCan and Trash-ICRA.} TrashCan~\cite{hong2020trashcan} and Trash-ICRA~\cite{ICRA2019} are datasets both containing underwater images. They are comprised of frames of video showing trash, remotely operated underwater vehicles (ROVs), and undersea flora and fauna. Both datasets are sourced from the JAMSTEC E-Library of Deep-sea Images (J-EDI) dataset [citation], curated by the Japan Agency of Marine-Earth Science and Technology (JAMSTEC) captured from real-world environments, providing a variety of objects. The clarity of the water and quality of the light vary significantly between images creating a diverse dataset. The image sizes in these datasets are 480x270px and 480x360px. Provided annotations are in COCO format. The TrashCan dataset is annotated on the instance segmentation level (7212 images and 6214 annotations) with 16 classes for Material Version (8 classes are trash related and followed \textit{trash}\_ name pattern) or 22 for Instance Version. On the other hand, the Trash-ICRA19 dataset is annotated on the detection level (7668 images and 6706 annotations). It contains seven categories based on the material of the objects.

\textbf{Drinking Waste.} Drinking Waste~\cite{drinking_waste_classification2020} contains over 4800 images of drinking waste belonging to 4 classes: Aluminium Cans, Glass bottles, PET bottles, and HDPE. Provided bounding-box annotations are in YOLO format. The dataset was created with a 12 MP phone camera. Images look similar -- there is usually one object in the center on the indoor, plain background. Most of the images have the size of 512x683px.

\textbf{MJU-Waste.} MJU-Waste dataset~\cite{MJU-Waste2020} is comprised of 2475 indoor trash images manually annotated in the form of an instance mask in COCO format. It allows two-class semantic segmentation (waste and background). For each color image, the co-registered depth image captured using an RGBD camera is provided. Objects are hand-held and mostly in the center of the image. In most cases there is only one object per image. The only image size is 640x480px.

\textbf{Cigarette butt.} The cigarette butt dataset~\cite{cigarette-butts} consists of above 2k images, which were synthetically composed photos of small cigarettes lying on the ground. It is an artificial dataset created by applying random scale, rotation or brightness to the foreground cutouts of 25 different cigarette butts, placed in around 300 ground photos. Instance mask annotations are provided for each cigarette object with a median size of 66x65px. Original resolution of images is equal 3024x4032px.

\textbf{Places.} Places~\cite{places} is a repository of 10 million scene photographs, labeled with 434 scene semantic categories, comprising a large and diverse list of the types of environments encountered in the world. Images were downloaded by online image search engines (Google Images, Bing
Images, and Flickr). Minimal size of images is 200x200px. Despite the fact that this is not a trash dataset, it can be used to identify natural and urbanized places without trash.

{
\begin{table}[!hbt]
\centering
\caption{Statistics for selected public waste datasets for classification and object detection purposes.}
\label{tab:dataset_stats}
\begin{tabular}{lllll}
\hline
\textbf{Dataset}         & \textbf{\# classes}     & \textbf{\# images}    & \textbf{\# instances}  & \textbf{annotation type} \\ \hline
TrashNet        & 5              & 2194         & 2194          & labels \\ 
Waste Pictures  & 34             & 23633        & 23633         & labels \\ 
Open Litter Map   & >100           & >100k         & >100k        & multilabels \\ \hline
Extended TACO    & 7            & 4562         & 14286          & bounding box \\
Wade-AI         & 1              & 1396         & 2247          & instance masks \\
UAVVaste        & 1              & 772          & 3718          & instance masks \\ 
TrashCan 1.0    & 8              & 7212         & 6214          & instance masks \\ 
Trash-ICRA19    & 7              & 7668         & 6706          & bounding box \\ 
Drinking waste  & 4              & 4810         & 5058          & bounding box \\ 
MJU-Waste       & 1              & 2475         & 2532          & instance masks \\  
Cigarette butt       & 1              & 2200         & 2200          & instance masks \\ \hline
Places       & 205              & 2.5M         & 2.5M          & labels \\  \hline
\end{tabular}
\end{table}
}

\section{Two-stage framework to detect and sort litter}
\label{sec:our_approach}
Litter sorting methods around the world differ significantly. On the one hand, this provides a large variety of objects (litter), as well as diverse backgrounds - waste can be commonly found indoor and outdoor, and in environments such as households, offices, roads and pavement scenes, and sometimes even under water. On the other hand, this affects available datasets, which do not provide a large enough number of annotated images. Moreover, a huge variety of categories and annotation levels (from classification to instance segmentation) resulted in the need for well-defined waste categories that could be applied in the industry. Hence, seven litter categories are provided based on a real-world waste segregation system in Gdańsk (Poland):  \textit{bio, glass, metal and plastic, non-recyclable}, \textit{other} (e.g. batteries, large household appliances, tires), \textit{paper}, and \textit{unknown} (old, degraded litter).

\begin{figure}[!ht]
\centering
  \includegraphics[width=0.85\textwidth]{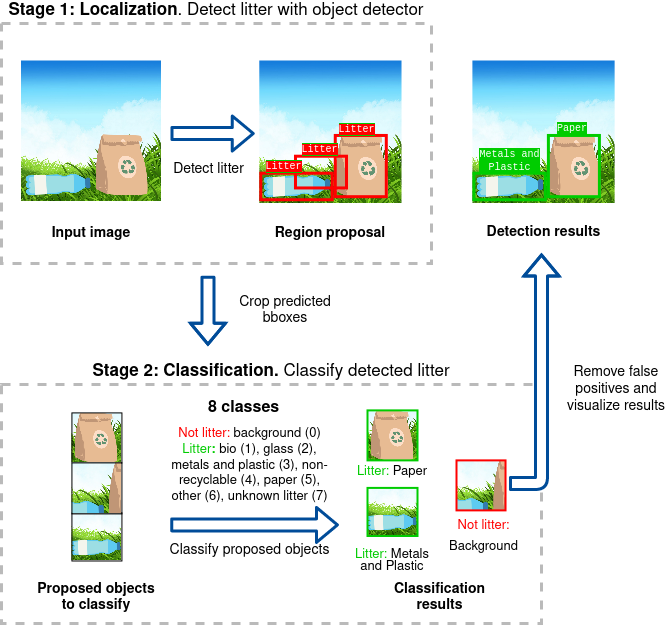}%
  \caption{A pipline of a two-stage framework to detect and sort litter\label{fig.idea2}}
\end{figure}

To utilize all of the available data we tackle the problem by dividing the detection into two separate steps: litter localization and litter classification. The localization model is used to find regions in the image containing garbage. Then, each region is extracted and passed to the classification model, that assigns its category. The proposed pipeline is presented in the Fig.~\ref{fig.idea2}.

This section outlines the details behind the training procedure, the methodology, and our experiments. We compare various architectures, as well as numerous hyperparameters, to ensure the efficiency of the solution. Additionally, we present how each dataset can be used in the training process - separately for waste localization, classification or both, depending on the annotation type.

\subsection{Litter in images}
\label{sec:datasets_collection}
During \textit{Detect Waste in Pomerania} project, twelve publicly available datasets and an additional data collected using Google Images Download~\cite{google-images-download2019} were used. Segmentation and detection datasets combined resulted in a new \textbf{detect-waste} dataset~\cite{proencca2020taco, UAVVaste2021, hong2020trashcan, ICRA2019, MJU-Waste2020, drinking_waste_classification2020, wade-ai2016} (with the addition of \textbf{cigarette butt} dataset~\cite{cigarette-butts} -- \textbf{detect-waste+} dataset) with a single class -- \textit{Litter} -- used in the first stage of our framework: \textbf{Localization} (see Fig.~\ref{fig.idea2} top panel). For the purpose of the second stage -- \textbf{Classification} -- additional images of \textit{bio, glass, other}, and \textit{paper}~\cite{TrashNet2016, waste_pictures2019, google-images-download2019} waste were added to the set of clipping images created by cutting out litter instances from some photos used in the detection stage~\cite{proencca2020taco, drinking_waste_classification2020, openlittermap2018}. These images form a final classification dataset named \textbf{classify-waste}. Furthermore, the dataset was supplemented with more than a thousand images containing garbage-free backgrounds~\cite{places} and around 55k pseudo-labeled images~\cite{openlittermap2018}, creating the \textbf{classify-waste+} dataset.

\begin{figure}[!ht]
\centering
  \includegraphics[width=0.8\textwidth]{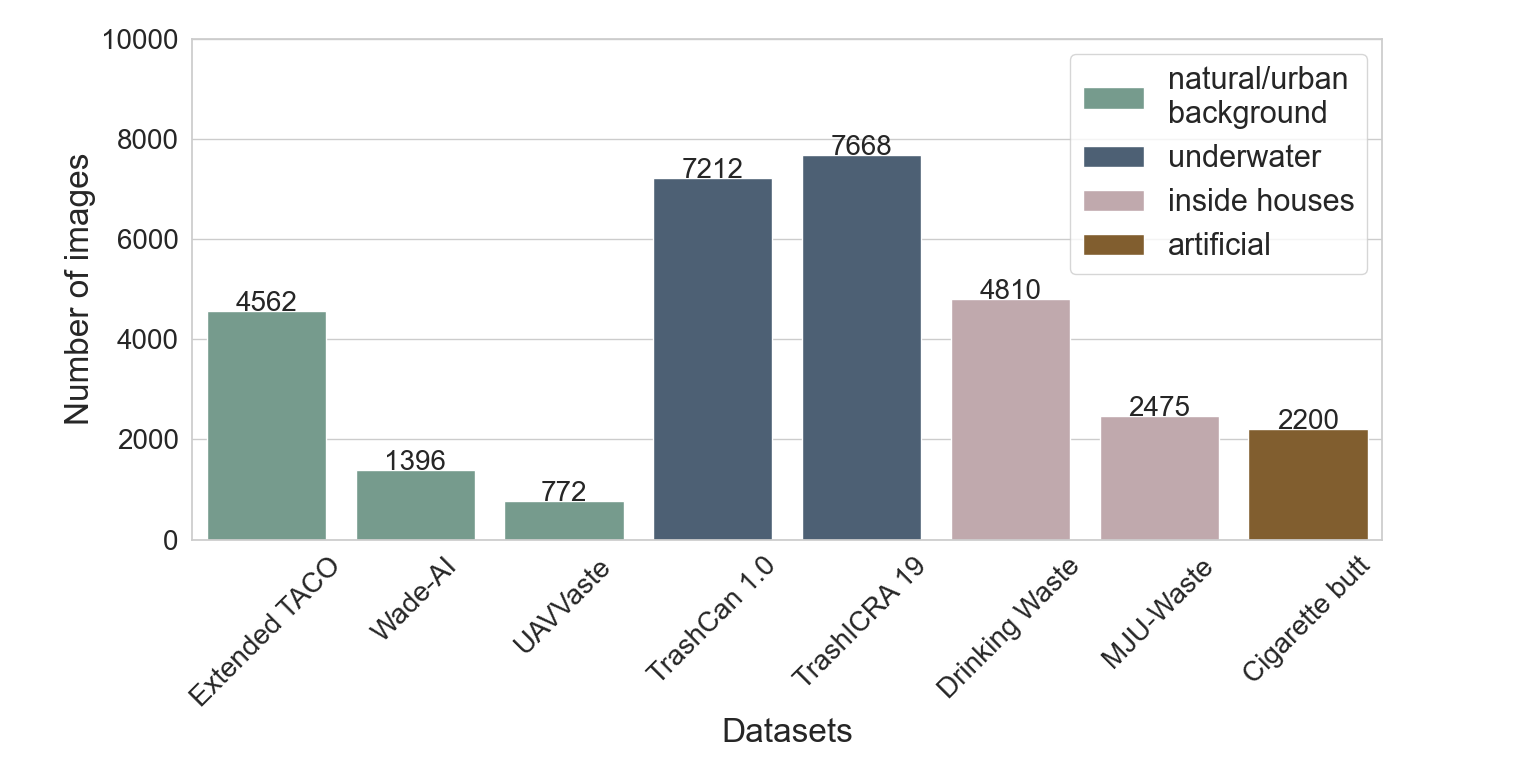}%
  \caption{Datasets included in \textbf{detect-waste+ dataset} used for detection task provided with number of images.\label{fig.mixedDataset}}
\end{figure}

\begin{figure}[!ht]
\centering
  \includegraphics[width=0.8\textwidth]{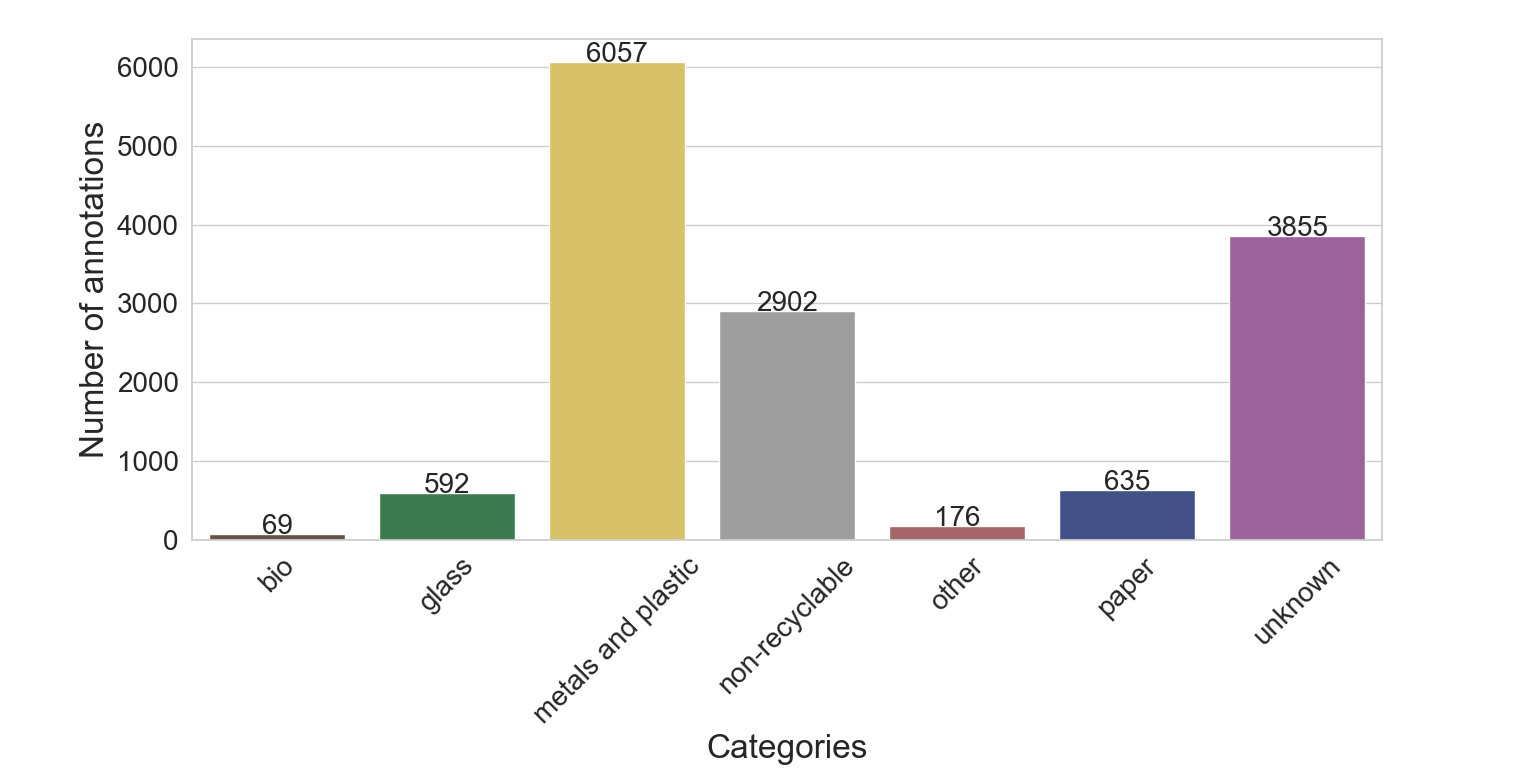}%
  \caption{Annotations per category for Extended TACO dataset.\label{fig.extendedTACO}}
\end{figure}

The localization task was mainly performed on a \textbf{detect-waste} dataset with a single class called \textit{litter}. In the last experiments, the dataset was extended to the \textbf{detect-waste+} dataset, which is previous \textbf{detect-waste} dataset with additional pictures presenting cigarette butts~\cite{cigarette-butts} to boost the detection of small objects. Fig.~\ref{fig.mixedDataset} shows the datasets included in the \textbf{detect-waste+} dataset along with the number of used images, divided into groups depending on the trash environment (in the wild, underwater, inside homes and artificial). Additional experiments were performed for multi-class detection on \textbf{Extended TACO}. The division of this dataset into Gdańsk segregation categories is presented in the Fig.~\ref{fig.extendedTACO}.

\begin{figure}[!hbt]
\centering
  \includegraphics[width=0.8\textwidth]{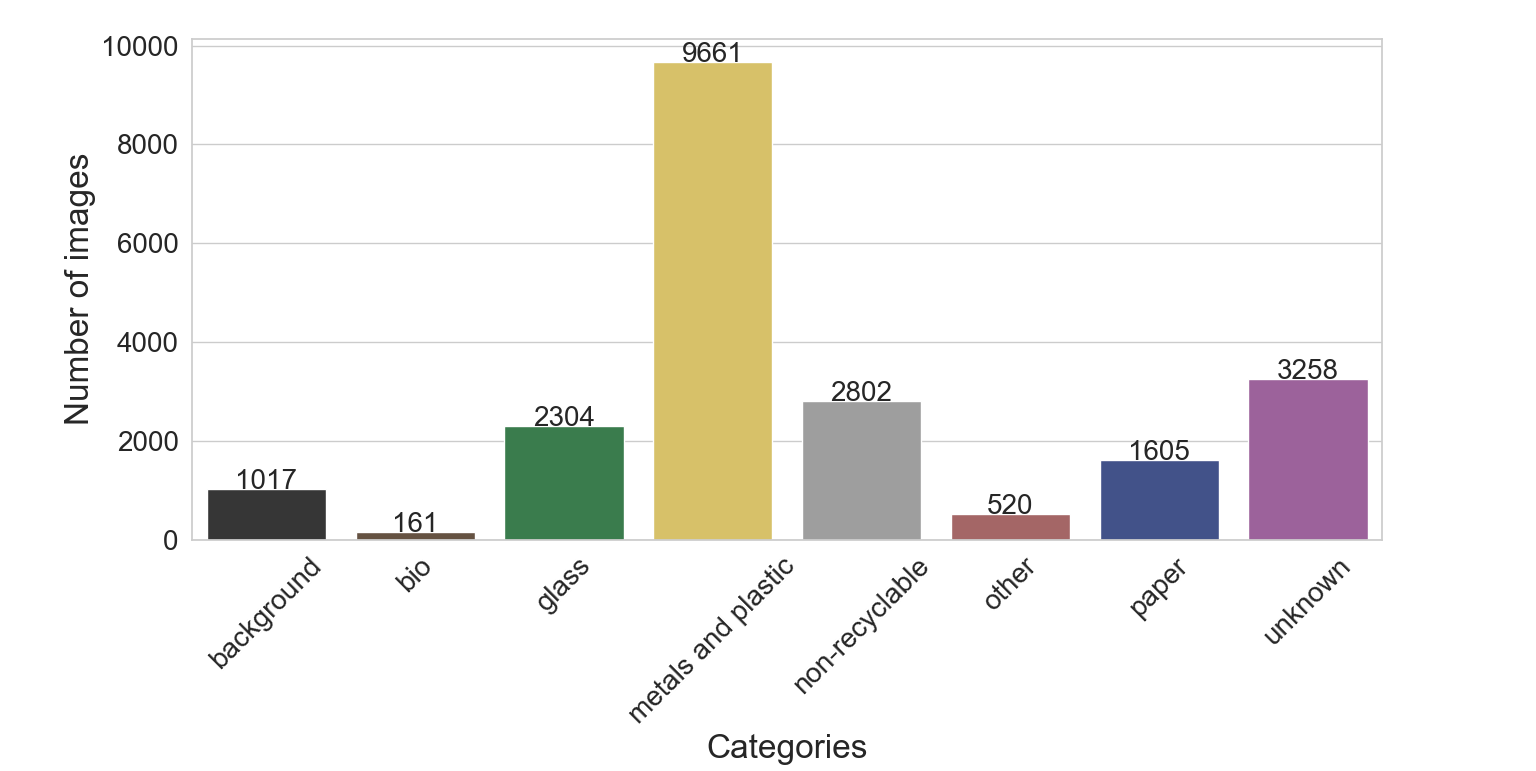}%
  \caption{Images per category for classify-waste + Places.\label{fig.imagesPerCategoryClassification}}
\end{figure}

In the classification task, in addition to instances cut-out from \textbf{Extended TACO} and \textbf{Drinking waste}, some classes from \textbf{TrashNet} (\textit{paper} and \textit{glass}) and \textbf{Waste Pictures} (\textit{bio} and \textit{other}) datasets have been added. Additionally, we scrapped remaining data from the web. The Google Images Download~\cite{google-images-download2019} software was used to search and collect more images with \textit{bio} and \textit{other} waste like batteries, medicines or rubble, and also images that depicted scenes without the presence of garbage (\textit{background}) from \textbf{Places} dataset~\cite{places}. Semi-labeled (with predicted localization) images from \textbf{Open Litter Map}~\cite{openlittermap2018} were also utilized. Fig.~\ref{fig.imagesPerCategoryClassification} shows number of used images per category (according to the obligatory segregation rules of the city of Gdańsk) without images from \textbf{Open Litter Map}, while Table \ref{tab:source_classification} provides information from which dataset the images come from, for which category. Table \ref{tab:pseudolabeling} presents numbers of images from semi-labeled Open Litter Map with some established seven litter class pre-assignment.

{
\begin{table}[!hbt]
\centering
\caption{Number of images and their origin for a given category for the classification task.}
\label{tab:source_classification}
\begin{tabular}{lllllllll}
\hline
\textbf{Dataset}    &\textbf{background}    &\textbf{bio}   &\textbf{glass} &\textbf{metals \& plastic}      &\textbf{paper} &\textbf{non-recyclable}  &\textbf{other} &\textbf{unknown} \\ \hline
Extended TACO   &-  &69  &592  &6057  &601  &2802  &154  &3258 \\
Drinking waste &-   &-  &1162   &3604   &-  &-  &-  &- \\
waste-pictures \& \\ Google search &-  &92 &49 &-  &203    &-  &366    &- \\
TrashNet    &-  &-  &501    &-  &801    &-  &-  &-\\
Places  &1017   &-  &-  &-  &-  &-  &-  &-\\ \hline
\end{tabular}
\end{table}
}

{
\begin{table}[!hbt]
\centering
\caption{Number of images from Open Litter Map with pre-assignment of litter class.}
\label{tab:pseudolabeling}
\begin{tabular}{ll}
\hline
\textbf{Category}         & \textbf{\# pseudo-labels}  \\ \hline
bio &- \\
glass  &3136   \\
metals and plastic  &29219  \\
non-recyclable &1971   \\
other  &900    \\
paper  &819    \\
unknown    &19701  \\ \hline
\end{tabular}
\end{table}
}

Most of the trash found in the classification dataset is \textit{metal and plastic}. Unfortunately, the second numerously represented category is \textit{unknown} -- the litter that has probably decomposed so much that it is hard to classify it. This makes our dataset highly imbalanced and requires special attention in the following steps. Another problem is related to annotations errors coming from datasets without strict annotation rules (especially when it comes to the trash label) -- unfortunately some annotations are of poor quality. Moreover, we manually rejected some of the images that were malformed or mislabeled.

To ensure that the training and test data distributions approximately match, we randomly selected 80\% of the original images as the training set. We kept the rest as validation/testing set for each part of the used dataset separately. The split was done by preserving the percentage of samples for each class.

\subsection{Litter Detection}
\label{sec:litter_detection}
In the first step, the proposed framework localizes litter in the image without recognizing its class. Three different neural networks popular in common object detection tasks were analyzed, namely EfficientDet~\cite{effdet-pytorch}, DETR~\cite{detr}, and Mask R-CNN~\cite{ren2015faster}. The first two architectures allow for object detection, whereas the third one also implements segmentation. All models were trained to predict trash at each location and scale (the solution is not targeted at a specific object size). To properly evaluate each proposed architecture, a set of tests was conducted using a wide range of datasets altogether and individually. Defined classes varied depending on the used dataset, as described in Section~\ref{sec:datasets_collection}.

\subsubsection{Training details}
To ensure one-stage detector efficient and fast performance, as EfficientDet~\cite{efficientdet2020}, was utilized to localize litter in images. The used model was initialized with pre-trained weights. In the final experiments decay rate was set to 0.95, the learning rate to 1e-3, and the number of epochs equaled 20. Also, several approaches to data augmentation~\cite{albumentations} were used. During training, Gaussian blur, random brightness, rain, fog, and snow were added. Additionally, images were rotated and cropped around the annotated bounding boxes with padding if needed. However, the best results were obtained for simple resize (to match default EfficientDet's input size), and normalization using means and standard deviation values per channel as for the COCO dataset~\cite{bib:COCO2014}. We hypothesize that augmentation did not significantly improve the results because used data was naturally diversified by mixing a wide range of datasets. In conducted studies, the Pytorch implementation of EfficientDet~\cite{effdet-pytorch}.

Also, Transformers approach was applied to the waste detection task. DETR (which stands for DEtection TRansformer) with ResNet-50 and ResNet-101 backbone was initialized with pre-trained weights. In all experiments AdamW~\cite{diederik2017adam} optimizer was used, with a learning rate set to 1e-4 in the transformer and 1e-5 in the backbone. Selected number of queries, that represents the maximum number of instances that could be found in image, equaled 100, as by default. In final experiments, after about 63 epochs loss values saturate, and longer training reduce neither the training nor the validation error. The DETR litter detection model is based on Facebook team implementation~\cite{detr}. 

In the case of instance segmentation, Mask R-CNN~\cite{he2017mask} with Resnet-50 backbone was used. In final experiments, the model was trained for 26 epochs with learning rate set to 1e-3 and decay set to 1e-4. The transfer learning technique was also used in this study, as each of the models was initially pre-trained on the COCO 2017~\cite{COCO2017} subset. Implementation of the model is based on standard Pytorch library. 

\subsubsection{Results}

Conducted experiments were divided into two phases. In the first phase selected architectures were tested on the \textbf{detect-waste} dataset. The model, which exhibit the best performance, was selected for further training. The second phase showed how the results are distributed on the subpart datasets. This allowed for a deeper analysis of the effectiveness of the selected architecture and better preparation to the final training. As a basic evaluation metric, Average Precision (AP) for intersection over union (IoU) equaled 0.50 was used (AP@05~\cite{pascal}). In the case of multi-label detection, this metric is averaged over all categories, giving mAP@0.50.

{
\begin{table}[!hbt]
\centering
\caption{Comparison of results of selected architectures  on \textbf{detect-waste} dataset with one class - \textit{litter}.}
\label{tab:archcomp}
\begin{tabular}{lll}
\hline
\textbf{Model}           & \textbf{Backbone}        & \textbf{AP@0.5} \\ \hline
DETR            & ResNet-50       & 50.7  \\ 
DETR            & ResNet-101      & 51.6  \\ 
Mask R-CNN      & ResNet-50       & 28.0  \\
\textbf{EfficientDet-D2} & \textbf{EfficientNet-B2} & \textbf{65.5}  \\ 
\hline
\end{tabular}
\end{table}
}

As presented in Table~\ref{tab:archcomp}, average precision of litter detection, using \textbf{detect-waste} dataset with one class, varied in the range of 28.0\% for Mask R-CNN with ResNet-50 backbone to 65.5\% for EfficientDet-D2. In general, due to a significant gap between EfficientDets and other tested architectures the rest of experiments were limited to this network family. Additionally, more complex networks from this family (i.e. EfficientDet-D3) were also tested, and exhibited a similar performance. As EfficientDet-D2 network reached the best evaluation results, it is smaller (considering the number of parameters) and requires less computing power, we decided to proceed with it exclusively.

\begin{figure}[!hbt]
\centering
  \includegraphics[width=\textwidth]{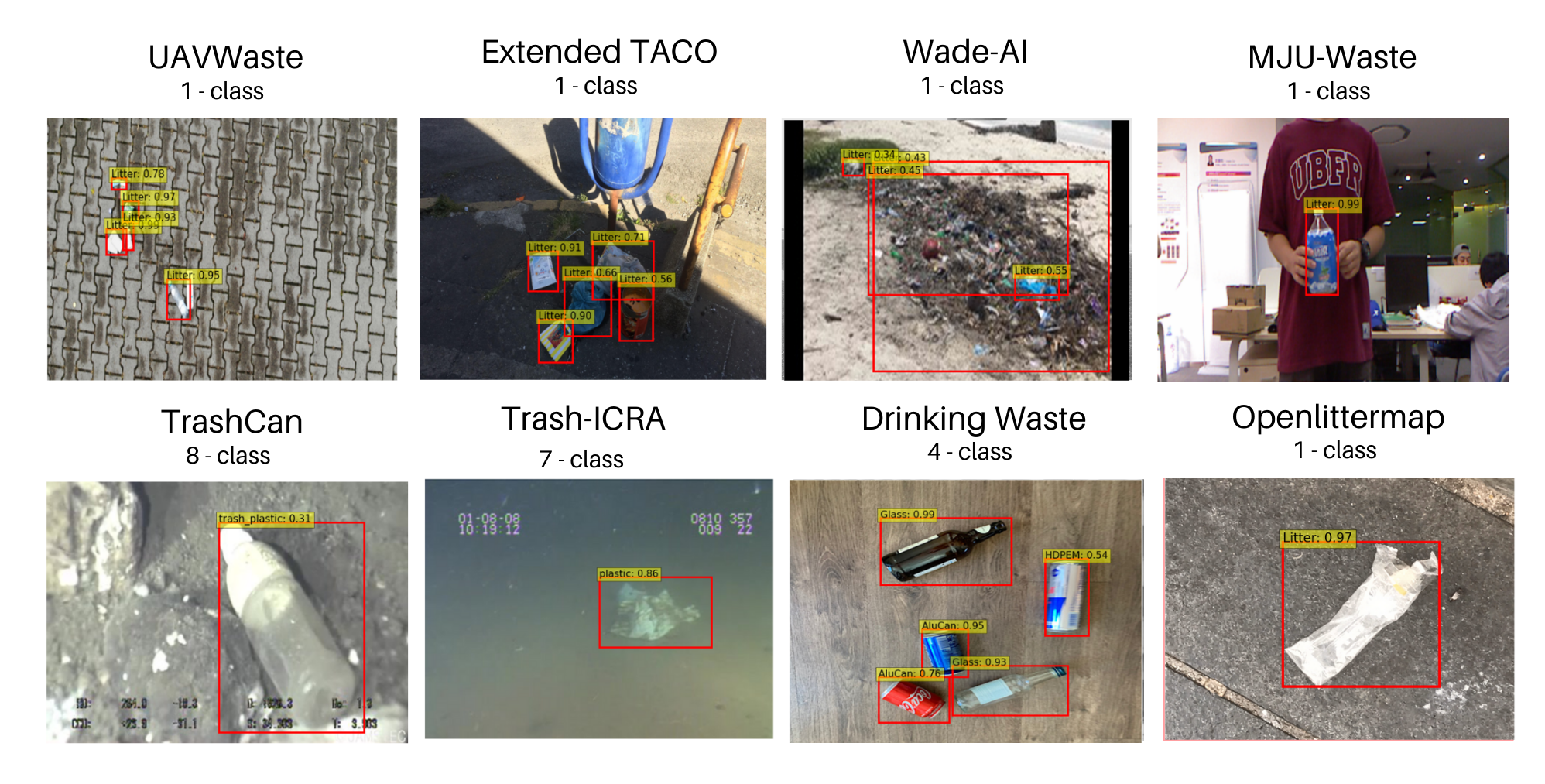}%
  \caption{Example EfficientDet-D2 predictions for the diverse waste datasets. Selected images were taken in different locations such as a beach, pavement, indoor, and underwater. Detected objects vary in size and number -- images show from one to five small, medium or large objects.\label{fig.detection_results}}
\end{figure}

\newpage
Results of a comprehensive study using EfficientDet-D2 and selected datasets separately are presented in Table~\ref{tab:datacomp}. The mAP@0.50 calculated per dataset is the highest for images presenting indoor scenario, namely \textbf{Drink-waste} (cans, plastic and glass bottles) and \textbf{MJU-Waste} (one hand-held waste object per image) datasets. However, EfficientDet-D2 also reached very high score, mAP@0.50 above 90\%, for \textbf{TrashCan 1.0} with selected 8 underwater waste categories. The worst result, reaching mAP@0.50 below 10\%, was achieved for the underwater images from \textbf{Trash-ICRA19} dataset, which could be related with poor quality of the photos (blurred movie frames). On the other hand, detection performance for images taken in natural or urban background was in range of 56.8\% for \textbf{Extended TACO} (trash in various environments) to 74.1 for \textbf{UAVVaste} (small objects constituting over 80\% of the dataset shown from a bird's eye view), which proved that precise detection of garbage in the different environment is possible. Corresponding sample predictions are shown in the Fig.~\ref{fig.detection_results}.

Achieved results proved importance of data quality in the learning process of DL-based system, but apart from different quality of photos and environments of waste occurrence in which they were taken, also the number and kind of waste classes varied depending on the analyzed dataset. Moreover, annotated trash objects differ in shape and size. This suggests that one-class detection (with one \textit{litter} category) using EfficientDet-B2 network might lead to much better performance. To provide more details, in final experiments with EfficientDet-D2, AP score at different IoU thresholds levels, along with AP@[0.50:0.95]~\cite{bib:COCO2014} (AP integrated over IoUs in range from 0.5 to 0.95, and step 0.05) were calculated. This results are presented in Table~\ref{tab:finalresults}. Selected neural network was tested in four different scenarios. In two of them the \textbf{Extended TACO} dataset was used, while the third and fourth experiments were conducted using different type of \textbf{detect-waste} dataset instead. 

{
\begin{table}[!hbt]
\centering
\caption{Results on different datasets achieved using EfficientDet-D2.}
\label{tab:datacomp}
\begin{tabular}{lllll}
\hline
\textbf{Dataset}       & \textbf{Classes} & \textbf{mAP@0.50} \\ \hline
Extended TACO          & 1                & 56.8  \\
Wade-AI                & 1                & 71.5  \\
\textbf{UAVVaste}      & 1                & \textbf{74.1}  \\ \hline
\textbf{TrashCan 1.0}  & \textbf{8}       & \textbf{91.3}  \\
Trash-ICRA19           & 7                & 7.3  \\ \hline
\textbf{Drink-waste}   & \textbf{4}       & \textbf{99.4}  \\
MJU-Waste              & 1                & 97.9  \\
\hline
\end{tabular}
\end{table}
}

{
\begin{table}[!hbt]
\centering
\caption{EfficientDet-D2 final evaluation results.}
\label{tab:finalresults}
\begin{tabular}{llllllll}
\hline
\textbf{Dataset}       & \textbf{Classes} & \textbf{mAP@0.50:0.95} & \textbf{mAP@0.50} & \textbf{mAP@0.75} & $\mathbf{AP_{S}}$ & $\mathbf{AP_{M}}$ & $\mathbf{AP_{L}}$\\ \hline
Extended TACO      & 7       & 11.9          & 16.2           & 13.0          & 6.4           & 9.4            & 15.0 \\
Extended TACO      & 1       & 40.4          & 56.8           & 43.5          & \textbf{19.8} & 37.2           & 51.6  \\
detect-waste       & 1       & 45.8          & 65.5           & 50.2          & 5.9           & 49.1           & 59.7  \\
detect-waste+      & 1       & \textbf{46.9} & \textbf{66.4}  & \textbf{51.3} & 9.3           & \textbf{51.3}  & \textbf{59.8}  \\
\hline
\end{tabular}
\end{table}
}

\newpage
EfficientDet-D2 trained on the \textbf{detect-waste+} dataset reached the biggest mAP for IoU equal both 0.50 (66.4\%) and 0.75 (51.3\%). Solution effectiveness in respect of detected object size demonstrated better precision for large (59.8\%) and medium (51.3\%) objects, than for small (9.3\%) instances. Moreover, as expected, the test conducted using the \textbf{detect-waste+} dataset, which boosted the evaluation result for small objects from 5.9\% to 9.3\%, and at the same time increased effectiveness in terms of AP@0.50 from 65.5\% to 66.4\%. In the case of a models trained on the \textbf{Extended TACO} dataset with one class, AP@0.50 reached 9.6 percentage points less (56.8\%) than the one that was trained on the \textbf{detect-waste+} dataset (66.4\%). However, in respect to detected objects size, among all conducted experiments, it reached the best results for small (19.8\%) objects.

On the other hand, experiments in which detected trash was divided into seven classes (\textit{bio, glass, metals and plastic, non-recyclable, other, paper, unknown}) resulted in significantly reduced mAP in all analyzed evaluation metrics, reaching almost 4 times smaller value for both IoU equal 0.50 (16.2\%) and 0.75 (13.0\%) than the best solution. Regarding detected objects size, for tiny objects multi-label detector trained on \textbf{Extended TACO} dataset reached better results (6.4\%) than the one that was trained on \textbf{detect-waste} dataset with one class (5.9\%). This may be due to the fact that in \textbf{Extended TACO}, approximately 45\% of instances are small (area < 32\textsuperscript{2}), while for \textbf{detect-waste} it is only almost 25\% of whole dataset. For that reason, feeding the \textbf{detect-waste} dataset with \textbf{cigarettes butt} improved the quality of prediction. It is worth emphasizing that a neural network that provides an ability to detect directly into 7 waste categories demonstrated average precision similar to one-class based detectors only for one category -- \textit{metals and plastic} -- reaching \mbox{AP@0.50 = 43.3\%}. Results obtained for the remaining six classes varied in range of \mbox{AP@0.50 = 0.1\%} for \textit{bio} to \mbox{AP@0.50 = 8.9\%} for \textit{unknown} litter. For that reason, it was decided to perform classification in a separate stage.

\subsection{Litter Classification}
\label{sec:litter_classification}
At the second stage of our approach, we performed multi-class classification for seven waste categories. Due to the imbalanced \textbf{classify-waste} dataset with a significant dominance of \textit{metals and plastic} class, classification networks were trained on cut out trash with some additions of underrepresented classes from other classification datasets as presented in Fig.~\ref{fig.imagesPerCategoryClassification} and described in Section~\ref{sec:datasets_collection}. Boundaries of cropped litter were established using bounding boxes for annotated images and objects detected by EfficientDet-D2~\cite{efficientdet2020} in case of unlabeled data. These all combined waste instances were applied as an input images to solve the classification problem.

\begin{figure}[!ht]
\centering
  \includegraphics[width=0.7\textwidth]{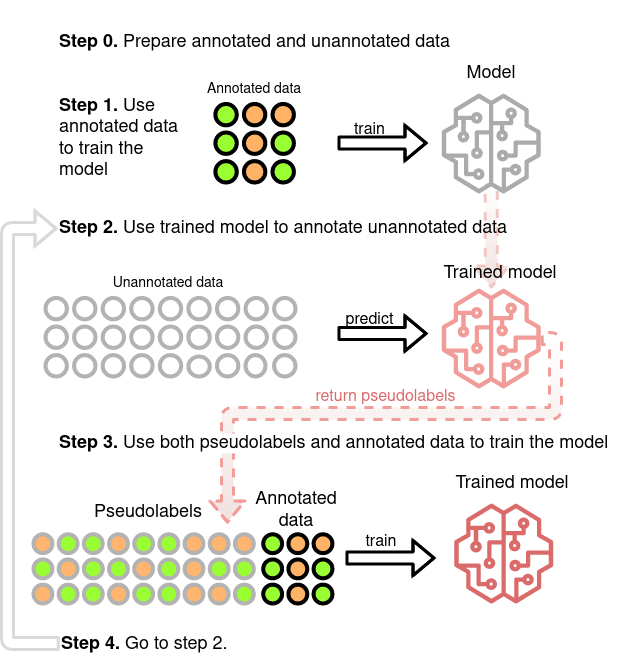}%
  \caption{Flowchart for pseudo-labeling semi-supervised learning technique.\label{fig.pseudolabel}}
\end{figure}

To take advantage of the semi-labelled \textbf{Open Litter Map} dataset~\cite{openlittermap2018} pseudo-labeling ~\cite{Lee2013PseudoLabelT} technique was applied. Pseudo-labeling is a kind of semi-supervised teaching method. The main concept of this approach is to teach a network to detect objects on data with and without annotation. Firstly unlabeled data is preassigned to some category by pretrained model, and afterwards used in further training. The process is repeated during the training every batch or epoch, as is presented in Fig.~\ref{fig.pseudolabel}. This provides considerably larger, yet only partially annotated, dataset. 

\subsubsection{Training details}

In our research we have tested two architectures dedicated to object classification problem - ResNet-50~\cite{resnet50} and EfficientNet-B2~\cite{efficientnet2019} to chose the one that achieves better performance for waste classification task. We have thoroughly examined number of training hyperparameters. During all experiments, we set the learning rate to 1e-4, batch size to 16, and trained each network for 20 epochs. The output of our network differed between 7 and 8 classes. The first 7 classes refer to previously mentioned waste categories, while the 8th was a \textit{background} class in order to reduce number of false-positives. Also, two kinds of samplers -- random and weighted -- were used to reduce the impact of data imbalance. Therefore it was crucial to investigate results on each class, not only on a whole dataset. The other hyperparameter was the "pseudo-labelling type" that shows if the pseudo-labels update was performed every batch or every epoch, or not at all. 

Moreover, to additionally ensure our training dataset's diversity, we applied data augmentation~\cite{albumentations}. At first, we extended the dimensions of our images to randomly crop parts of them. Horizontal/vertical flip and shift scale rotation, which does not change the image's content but only the position of an object in the image, were also applied. On the other hand, a random brightness contrast and cutout were used to change the image's content slightly. Finally, both training and testing sets were normalized and resized to 224x224, as it was an EfficientNet-B2~\cite{efficientnet2019} input size. The second stage of our approach was self-implemented in Pytorch Lightning and based on the repository~\cite{EfficientNet-git}.

\subsubsection{Results}

A comparison of the performance of two different neural networks is presented in Table~\ref{tab:classifiers_comp}. One of them is EfficientNet-B2 - the backbone of the network used in the detection task. The results achieved using this classifier exceeded markedly higher values (over 10 percentage points) than ResNet-50. EfficientDet-B2 being state-of-the-art architecture comparing to ResNet-50 provides better performance in the classification task, similary as in detection, where backbones are EfficientNet-B2 for EfficientDet-D2 and ResNet-50 for Mask RCNN or DETR (see Table~\ref{tab:archcomp}). It leads to the conclusion, that the EfficientNet networks family is a preferred choice to deal with waste images.

The experiments showed that updating pseudo-labels every batch can slightly raise accuracy. Although the best-achieved accuracy was 74.6\%, it was performed for a random sampler. While analyzing the confusion matrices of each training, we have noticed that applying a weighted sampler provides more balanced results for each class. Therefore we have achieved an accuracy of 73\% (and 86.7\% on the training set), while almost 25\% of our dataset were test images.

\begin{figure}[htb]
    \begin{minipage}[c]{.55\textwidth}
        \centering
        \includegraphics[width=\textwidth]{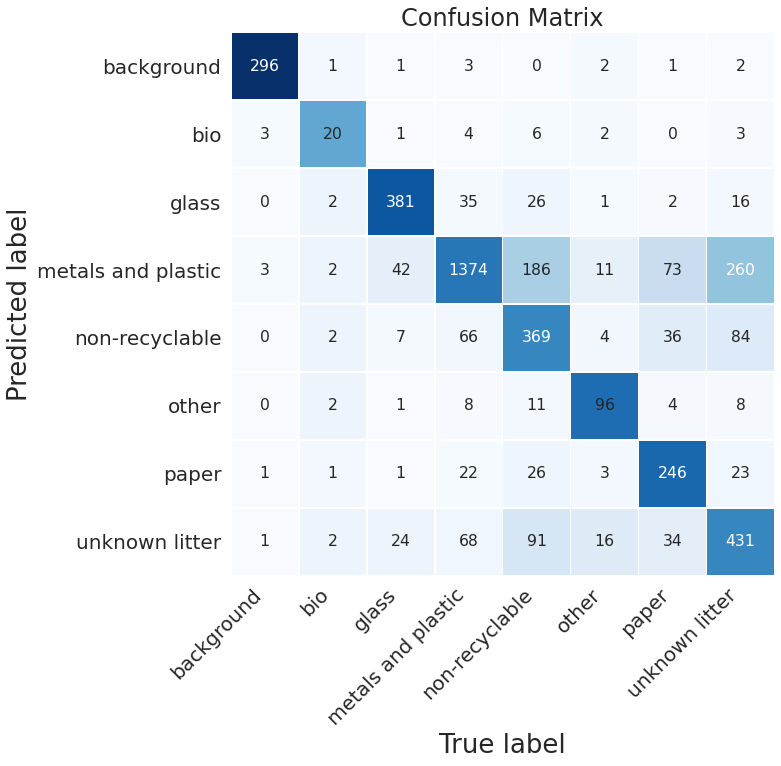}
    \end{minipage}
    \hfill
    \begin{minipage}[c]{.35\textwidth}
        \centering
        \includegraphics[width=\textwidth]{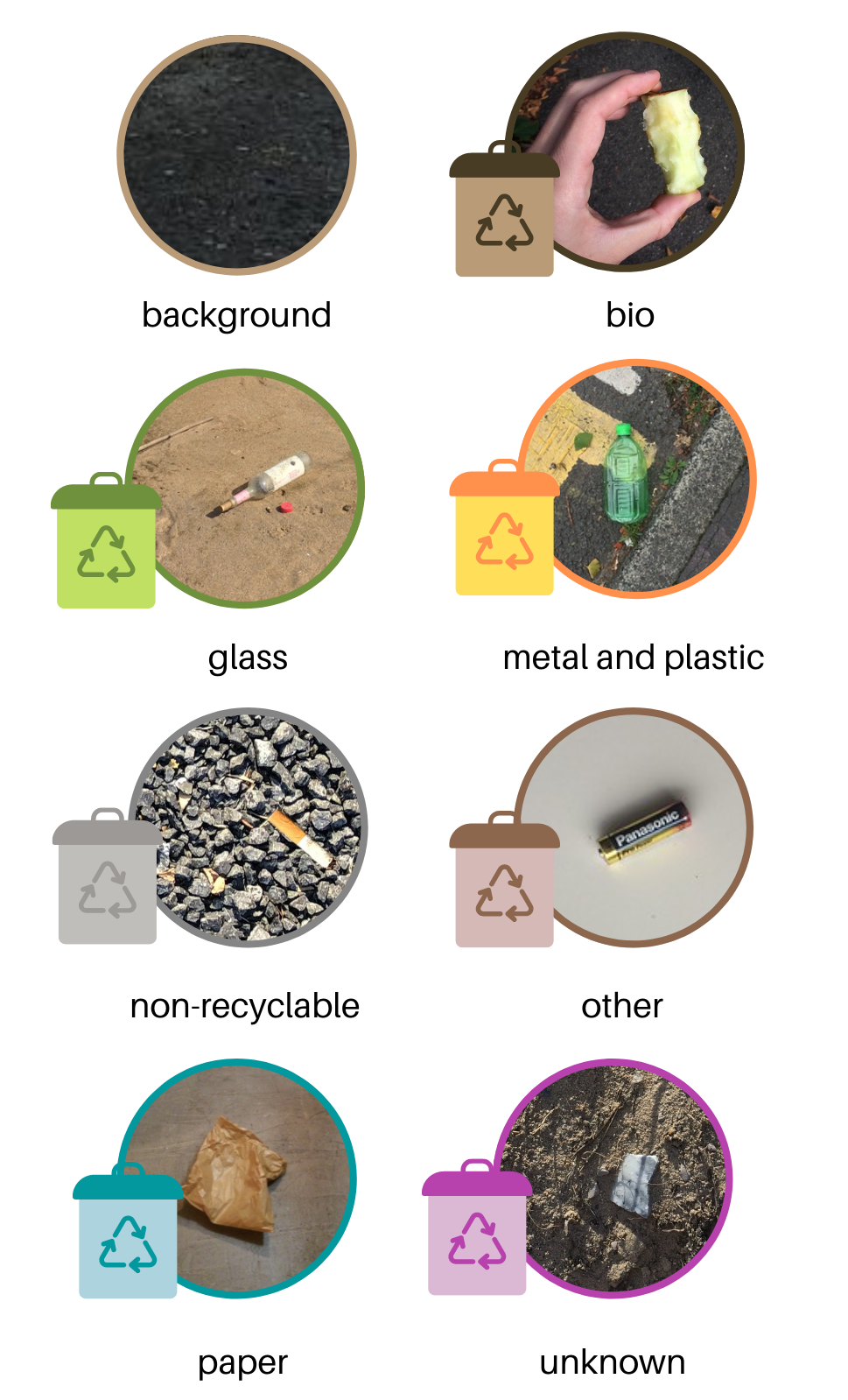}
    \end{minipage}
    \caption{Evaluation of the accuracy of a classification in form of confusion matrix for EfficientNet-B2, Weighted Sampler, and Pseudolabelling per batch. \label{fig:conf_matrix}}
\end{figure}

Although the confusion matrix clearly shows that most of the predictions are accurate (see Fig.~\ref{fig:conf_matrix}, and Table~\ref{tab:classifcation_report}), it indicates a significant data imbalance. The \textit{metals and plastic} class was predicted with the highest precision of 87\%, which is connected to the large representatives of this class. Still, it also results in a relatively low recall, which means that many objects are classified incorrectly as \textit{metals and plastic}. There was a noticeable problem with identifying the \textit{unknown} and \textit{non-recyclable} classes, which precision was equal 52\%. The other classes were recognized with the higher precision but still, due to the data imbalance, not fully correct.

The biggest confusion was remarked between \textit{metals and plastic} and \textit{unknown} category. Probably it was because of partly degraded or destroyed trash. All the classes were rarely misclassified with the \textit{glass}, as evidenced by a high recall value - 82\%. The F$_1$-score metric for this class achieved the utmost result - 83\%, which shows the balance between false positives and false negatives. It is worth noticing, that eliminating background from the rest of the waste was extremely successful - at the level of 97\% for precision and 97\% for recall. Adding a separate class for background improved the performance and as it was assumed, reduced the number of false-positives.

{
\begin{table}[!hbt]
\centering
\caption{Comparison of accuracy of selected classifiers.}
\label{tab:classifiers_comp}
\begin{tabular}{lllll}
    \hline
        \textbf{Model} & \textbf{\# classes} & \textbf{Accuracy} & \textbf{Sampler} & \textbf{Pseudo-labeling type} \\ \hline
        \textbf{EfficientNet-B2} & \textbf{8} & \textbf{73.02} & \textbf{Weighted} & \textbf{per batch} \\
        EfficientNet-B2 & 8 & 74.6 & Random & per epoch \\
        EfficientNet-B2 & 8 & 72.8 & Weighted & per epoch \\
        EfficientNet-B4 & 7 & 71.0 & Random & per epoch \\
        EfficientNet-B4 & 7 & 67.6 & Weighted & per epoch \\
        EfficientNet-B2 & 7 & 72.7 & Random & per epoch \\
        EfficientNet-B2 & 7 & 68.3 & Weighted & per epoch \\
        EfficientNet-B2 & 7 & 74.4 & Random & None \\
        ResNet-50 & 8 & 60.6 & Weighted & None \\ \hline
    \end{tabular}
\end{table}
}

\begin{table}[!ht]
    \centering
    \caption{Summary of the precision, recall, and F$_1$-score for each class of waste.}
    \label{tab:classifcation_report}
    \begin{tabular}{llllll}
    \hline
        \textbf{Class name} & \textbf{Precision} & \textbf{Recall} & \textbf{F$_1$-score} \\ \hline
            \textit{background} & 0.97 & 0.97 & 0.97 \\
            \textit{bio} & 0.62 & 0.51 & 0.56 \\
            \textit{glass} & 0.83 & 0.82 & 0.83 \\
            \textit{metals and plastic} & 0.87 & 0.70 & 0.78 \\
            \textit{non-recyclable} & 0.52 & 0.65 & 0.58 \\
            \textit{other} & 0.71 & 0.74 & 0.72 \\
            \textit{paper} & 0.62 & 0.76 & 0.68 \\
            \textit{unknown litter} & 0.52 & 0.65 & 0.58 \\ \hline
    \end{tabular}
\end{table}

\newpage
\section{Conclusions and Future Work}
\label{sec:conclusions}

There is a visibly increasing demand for artificial intelligence in numerous human activities. Following that, we proposed a DL-based framework that can localize trash in the image and then identify its class using two separate neural networks. Datasets used in this study were created by using various publicly available data of waste collected in diverse environments: inside houses, in the natural or urban environment and even underwater.

Firstly, three neural networks architectures were adapted to conduct the garbage detection. Used data was naturally augmented, which allowed for a precise location of the waste with AP@0.5 equals 66.4\% for EfficientDet-D2 model. This is an excellent result comparing to other recent reports, that were presented for some of used datasets separately (15.9\% for TACO with Mask R-CNN~\cite{proencca2020taco}, or 55.4\% for TrashCan 1.0 with Mask R-CNN~\cite{hong2020trashcan}). Qualitatively good results were also observed for the semi-labeled OpenLitterMap dataset, allowing for its further use in the next stage.

In the case of classification, \textit{litter} was divided into seven categories that imitate the sorting policy introduced in the city of Gdańsk with extra \textit{background} class to eliminate false positives coming from the detector. Additionally, during training pseudo-labeling technique was applied, which allowed to utilize an unlabeled data. However, this gave only a slight performance boost in case of litter classification, which can be related to high imbalance and small amounts of labeled data for specific litter categories, especially in the case of \textit{bio} or \textit{other} waste. In the end, the accuracy up to 75\% was achieved for the EfficientNet-B2 network. To the best of our knowledge, we present one of the first results to classify litter in wild.

As Artificial Intelligence is required to be more accurate than a human, the main future direction for the proposed system will be to improve its performance. Selected detectors works well when localizing medium and large objects, but recognition of small litter is still challenging. For that reason, exploring different state-of-the-art approaches, such as Deformable DETR~\cite{zhu2020deformable}, seems to be a good idea. On the other hand, a more balanced dataset and the use of the latest EfficientNetv2~\cite{tan2021efficientnetv2}, could also boost the classification accuracy.

Despite this, the presented framework showed the great potential of the DL-based methodology for waste management. In the future with the assistance of DL models, it would be possible to mount robotic arms in waste management plants to automatically distinguish between different classes of objects and sort garbage without human intervention. Additionally, the high precision of litter localization in a large variety of the environments, shows the possibility of using neural networks for waste monitoring in cities or detecting of illegal dumps in nature, for example with the use of drones.

\section{Acknowledgements}

MF, ZK, SM and AM contributed to the final version open-source code. MF, ZK and SM analyzed and interpreted the waste datasets regarding the waste sorting rules in Gdańsk (Poland). AM and SM planned the experiments and adjusted the models. SM prepared the results of the experiments. AM lead the team during the project. SM took the lead in the preparation of the manuscript. AM prepared the illustrations. AK and KM helped supervise the project.

SM wrote the Introduction. SM and AM conducted the researched on available datasets. AK conducted and written the review on the deep learning classification. KM conducted and written the review on the deep learning object detection. ZK analyzed the data and written the datasets sections. MP and MF described our approach and experiments. SM concluded our project and proposed future directions.

All authors provided critical feedback and helped shape the research, analysis and manuscript.

The 5-month (October 2020 -February 2021) lasting project "Detect Waste in Pomerania" was organized and lead by Agnieszka Mikołajczyk, Magdalena Kortas and Ewa Marczewska from Women in Machine Learning \& Data Science Trojmiasto. A team of carefully selected members, nine female data scientists, analysts and machine learning engineers supported by five industry mentors studied and worked together on developing a model for trash detection. This solution would be applicable for video and photography. The authors acknowledge other team members: Anna Brodecka, Katarzyna Łagocka, Ewa Marczewska, Pedro F. Proença, Adam Kaczmarek and Iwona Sobieraj who contributed to the project.

The authors acknowledge the infrastructure and support of Digital Innovation Hub dih4.ai in Gdańsk. The authors wish to express their thanks to Voicelab.ai for the financial support and Epinote for data annotation.

\medskip
\bibliographystyle{unsrt}
\bibliography{references}

\end{document}